\documentclass[smallabstract,smallcaptions]{dccpaper}

\usepackage{epsfig}
\usepackage{citesort}
\usepackage{amsmath}
\usepackage{amssymb}
\usepackage{color}
\usepackage{soul}
\usepackage{url}

\usepackage[utf8]{inputenc}
\usepackage{wrapfig}
\usepackage[normalem]{ulem}
 \useunder{\uline}{\ul}{}
 \usepackage[table,xcdraw]{xcolor}
\usepackage{subcaption}
\usepackage{graphicx}
\usepackage{float}
\usepackage{adjustbox}
\usepackage{xcolor}
\usepackage{enumitem}
\usepackage{booktabs}
\usepackage{multirow}
\usepackage[ruled,vlined]{algorithm2e}
\usepackage{pifont}

\newlength{\figurewidth}
\newlength{\smallfigurewidth}

\begin{document}

\title
{\large
\textbf{AIDetx: a compression-based method for\\ identification of machine-learning generated text}
}

\author{%
Leonardo Almeida$^{2}$, Pedro Rodrigues$^{2}$, Diogo Magalhães$^{2}$,\\ Armando J. Pinho$^{1,2}$, Diogo Pratas$^{1,2,3}$\\[0.5em]
{\small\begin{minipage}{\linewidth}\begin{center}
\begin{tabular}{ccc}
$^1$IEETA/LASI - Institute of Electronics and Informatics Engineering of Aveiro\\
$^2$DETI - Department of Electronics, Telecommunications and Informatics\\
$^3$DoV - Department of Virology, University of Helsinki, 00014 Helsinki, Finland\\
University of Aveiro, 3810-193 Aveiro, Portugal\\ \url{{leonardoalmeida7,pedrofrodrigues4 ,d.magalhaes,ap,pratas}@ua.pt} 
\end{tabular}
\end{center}\end{minipage}}
}

\maketitle
\thispagestyle{empty}

\begin{abstract}
This paper introduces AIDetx, a novel method for detecting machine-generated text using data compression techniques. Traditional approaches, such as deep learning classifiers, often suffer from high computational costs and limited interpretability. To address these limitations, we propose a compression-based classification framework that leverages finite-context models (FCMs). AIDetx constructs distinct compression models for human-written and AI-generated text, classifying new inputs based on which model achieves a higher compression ratio.
We evaluated AIDetx on two benchmark datasets, achieving F1 scores exceeding 97\% and 99\%, respectively, highlighting its high accuracy. Compared to current methods, such as large language models (LLMs), AIDetx offers a more interpretable and computationally efficient solution, significantly reducing both training time and hardware requirements (e.g., no GPUs needed). The full implementation is publicly available at \url{https://github.com/AIDetx/AIDetx}.
\end{abstract}

\Section{Introduction}
Recently, the advancement of artificial intelligence (AI) has revolutionised numerous industries. Healthcare, aviation, agriculture, and financial services have all undergone big changes with the rise of AI. Although most applications of AI have a positive impact on society, there are concerns about the potential misuse of this technology. One of the main concerns is related to AI-generated content, which can be in various forms, from news articles and social media to creative works such as photography or design. The misuse of AI-generated content can have a serious impact on society; for example, spread misinformation and manipulate public opinion \cite{waldrop2017genuine}.

To face this challenge, the research community has been working on developing methods capable of distinguishing between human-generated and AI-generated content, more specifically, text. Currently, the most popular approach to this problem is to train deep learning models on large datasets of human and AI-generated text to learn how to differentiate between the two. Known practical examples of this approach are the GPTZero and OpenAI Classifier tools. Although these models have shown promising results, they suffer from several limitations, including high computational costs \cite{Angelov2016ChallengesID}, issues with interpretability and explainability \cite{DLChallenges}, the need for substantial input text to produce reliable outcomes \cite{Majovsky2024}, among others.

An alternative approach to the text classification problem comes from the field of information theory, specifically through the use of data compression techniques. Data compression leverages the statistical structure of information to reduce its size, encoding frequently occurring patterns more efficiently. Over the past few decades, the research community has demonstrated the success of applying data compression techniques to classification problems. For example, Khmelev \cite{KukushkinaPK01} performed experiments using a large variety of compression methods for author classification. Similarly, Benedetto et al. \cite{Benedetto_2002} demonstrated how compression algorithms like gzip can be used to classify written texts in different languages based on their compressibility. Pinho et al. \cite{pinho2018application} applied data compression for handwritten digit classification. More recent works, such as that of Saikrishna et al. \cite{7893212} and Nishida et al. \cite{10.1145/2064448.2064473}, have explored the application of data compression techniques for spam filtering and tweet classification. 

In this paper, we explore the usage of data compression to classify a text as human-written or AI-written. To represent data dependencies, we rely on finite-context models, which are a specific type of Markov models \cite{bell1990text,sayood2017introduction,pinho2010finite}.

The remaining sections of this paper are organised as follows: Section 2 presents the methodology; Section 3 describes the experiments carried out to benchmark the method; Section 4 and 5 presents the obtained results and conclusions, respectively.

\Section{Methodology}\label{chap:methodology}

To classify a text using data compression, the core idea involves building a model or dictionary for each class, by compressing files that represent those classes. This compression process acts as a form of ``training'' the classifiers on labelled documents from each class. When a new document is introduced, it is assigned to the class that results in the highest compression ratio.

From an information-theoretic perspective, the compression ratio reflects the cross-entropy between the training text and the new document. The document is assigned to the class whose training text minimises this cross-entropy, indicating a better fit between the new document and that class's model.

\begin{figure}[ht]
    \centering
    \includegraphics[width=.86\linewidth]{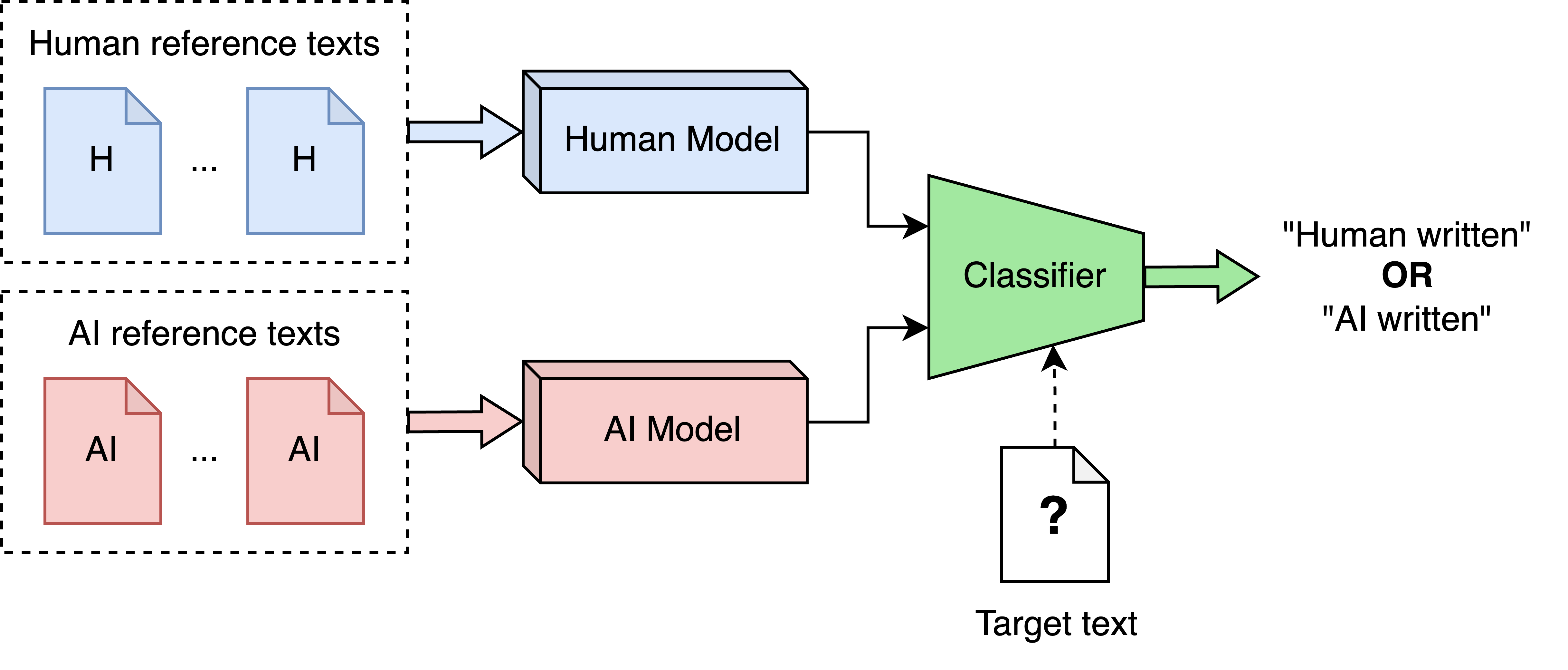}
    \caption{Overview of the classifier based on finite-context models (FCMs).}
    \label{fig:methodology}
\end{figure}

In simpler terms, the approach works as follows: for each class, represented by a reference text $r_i$, we create a model that is a good description of $r_i$. By a “good description” we mean a model that requires fewer bits to describe $r_i$ than other models, or, in other words, that is a good compression model for the “members of the class” $r_i$. Then, we assign to the target text $t$ the class corresponding to the model that requires fewer bits to describe it, i.e., to compress $t$.
A simple illustration of this method is presented in Fig. \ref{fig:methodology}.

For building a model that represents each class, we used the concept of finite-context models (FCM). A FCM is a probabilistic model relying on the Markov property, which provides the probability of the next symbol given a certain context depth. With these probabilities, it is then possible to determine the number of bits required to encode a symbol, which is fundamental to determine which model presents higher compression ratio on the target text.

Specifically, the number of bits needed to represent a target string is given by

\begin{equation}
    \sum_{i=1}^{n} -\log_{2}P(x_{i}|x_{i-1}, x_{i-2}, ..., x_{i-k}),
\end{equation}
where $x_{i}$ is the symbol at position \textit{i} in the text, and \textit{n} is the length of the text. On the other hand, the probability of a symbol to appear in a given context is given by

\begin{equation}
    P(x_{i}|x_{i-1}, x_{i-2}, ..., x_{i-k}) = \frac{N(x_{i-1}, x_{i-2}, ..., x_{i-k}, x_{i}) + \alpha}{\sum_{j\in\Sigma} N(x_{i-1}, x_{i-2}, ..., x_{i-k}, j) + \alpha |\Sigma|},
\end{equation}
where $N(\cdot)$ is the number of times a certain sequence of symbols appears in the model, $\alpha$ is the smoothing factor, and $\Sigma$ is the alphabet.

As can be seen in the formulas, there are three main parameters that need to be optimised: $\alpha$, $k$ and $\Sigma$. The optimisation of these parameters is reported in the following section.

The method was implemented into a computer tool, called AIDetx, using the C++ and Python languages, and is available, under GPL V3 license, at \url{https://github.com/AIDetx/AIDetx}. AIDetx provides a command-line interface that allows the user to train the models, classify the target texts, and evaluate the classifier's performance.

\Section{Benchmark}

To benchmark the method, we used a dataset constituted by AI-generated and human-generated text. Subsequently, parameter optimisation was performed, followed by an evaluation of the impact resulting from trimming the alphabet.

\subsection*{Datasets}

To accurately create models, high-quality datasets are critical. We used two public datasets from the Hugging Face Datasets repository: HC3\footnote{\url{https://huggingface.co/datasets/Hello-SimpleAI/HC3}} with 147 MB and AI-human-text\footnote{\url{https://huggingface.co/datasets/andythetechnerd03/AI-human-text}} with 571 MB. The HC3 dataset, introduced in \cite{guo2023closechatgpthumanexperts}, contains 24,321 samples, and features human and ChatGPT-generated answers to various questions. We pre-processed this dataset by extracting answers, removing duplicates and short samples, and balancing character counts between classes. The AI-human-text dataset has around 400,000 samples, featuring human and AI texts labelled accordingly. Similar pre-processing was performed to ensure balanced representation.

We divided the datasets into training (80\%), validation (10\%), and test sets (10\%). The training set is used to build the models, the validation set to optimise the hyperparameters, and the test set to assess classifier's performance. The compiled samples serve as reference texts for model training.

\subsection*{Parameter Optimisation}

For the finite-context models to be most effective in distinguishing between human and AI generated text and have good time performance, we need to study the best values for the hyperparameters that will be used to generate and to use the models.

In the Methodology section, we concluded that three hyperparameters can be fine-tuned: $k$, the order of the Markov model; $\alpha$, the smoothing factor; $\Sigma$, the alphabet used by the models to generate the tables for classifying the target texts. The first two hyperparameters are used to create the models, and the last one is used to classify the target texts.

To determine the best values for the hyperparameters, we performed an exhaustive search, varying the values of each parameter. We then analysed the model's performance in the validation samples using the F1 score (as the number of samples for each type was unbalanced) and the time performance as metrics. To achieve this, we followed a systematic approach where first we tested a combination of different values for the hyperparameters \textit{k} and $\alpha$ using grid search and then, already using the selected values, we tested different values for the alphabet.
The values tested for the first step were: $k=\{3, 4, 5, 6, 7, 8, 9, 10\}$ and $\alpha = \{0.1, 0.5, 1, 5, 10\}$. 

\begin{figure}[ht]
    \centering
    \begin{subfigure}[b]{0.49\linewidth}
        \includegraphics[width=\linewidth]{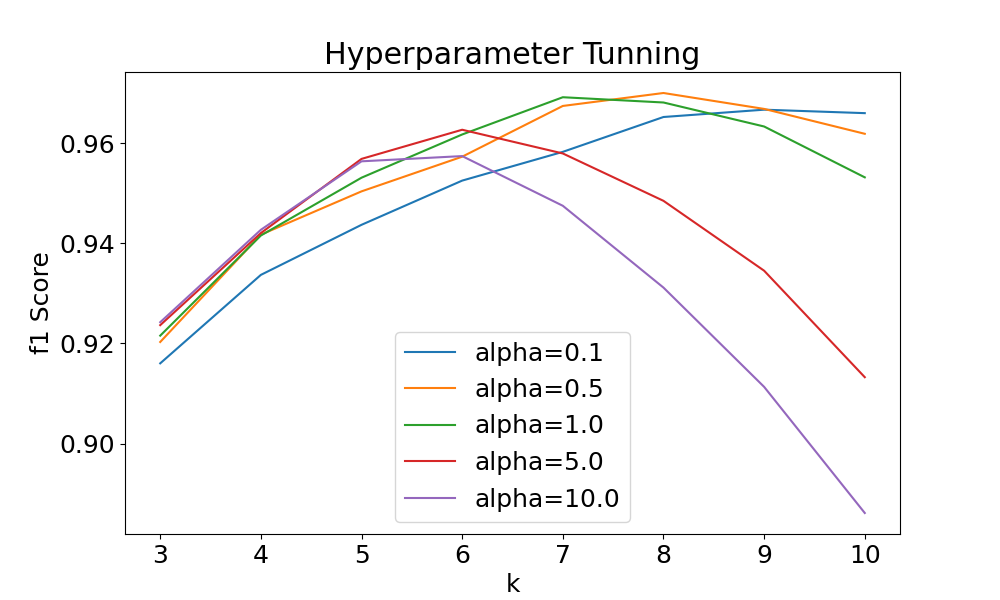}
        \captionsetup{width=.9\linewidth}
        \caption{HC3}
        \label{fig:hyperparameters2}
    \end{subfigure}
    \begin{subfigure}[b]{0.49\linewidth}
        \includegraphics[width=\linewidth]{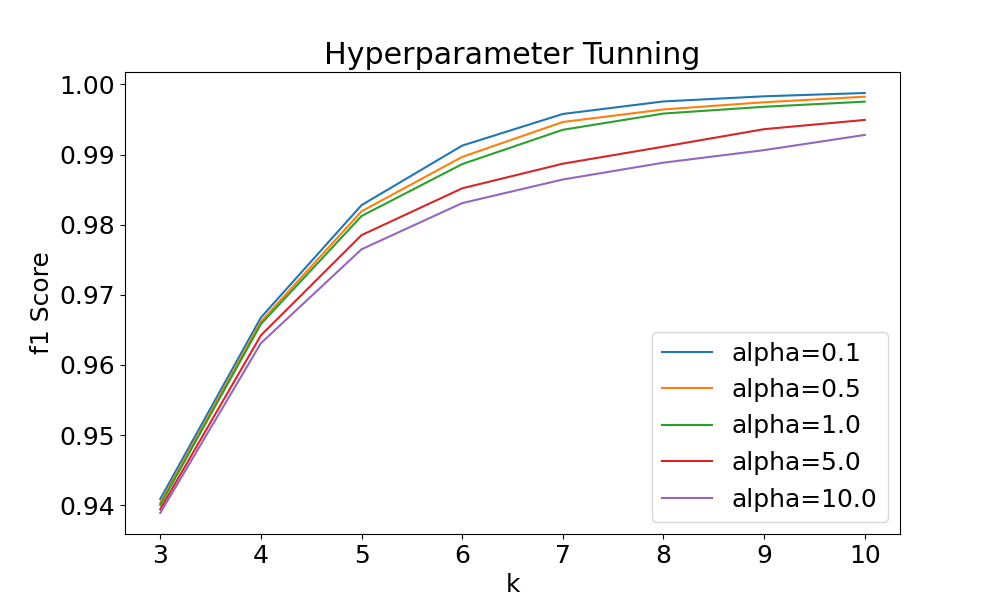}
        \captionsetup{width=.9\linewidth}
        \caption{AI-human-text}
        \label{fig:hyperparameters1}
    \end{subfigure}
    \caption{F1 score for the grid search of the hyperparameters $k$ and $\alpha$ for the datasets HC3 (On the left) and AI-human-text (On the right).}
    \label{fig:hyperparameters}
\end{figure}

\begin{figure}[H]
    \centering
    \begin{subfigure}[b]{0.49\linewidth}
        \includegraphics[width=\linewidth]{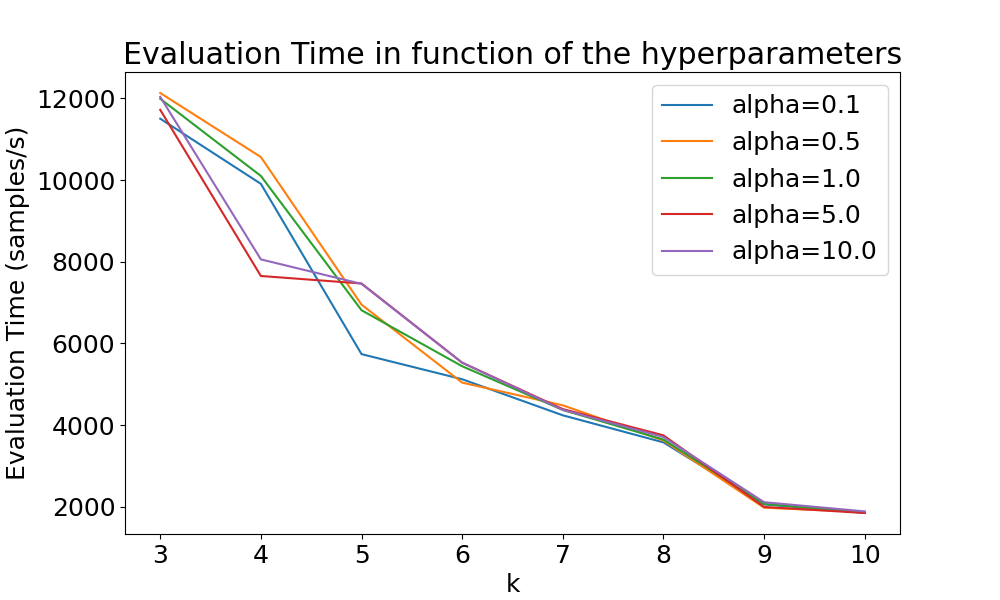}
        \captionsetup{width=.9\linewidth}
        \caption{HC3 (858 chars/samples)}
        \label{fig:hyperparameters_time2}
    \end{subfigure}
    \begin{subfigure}[b]{0.49\linewidth}
        \includegraphics[width=\linewidth]{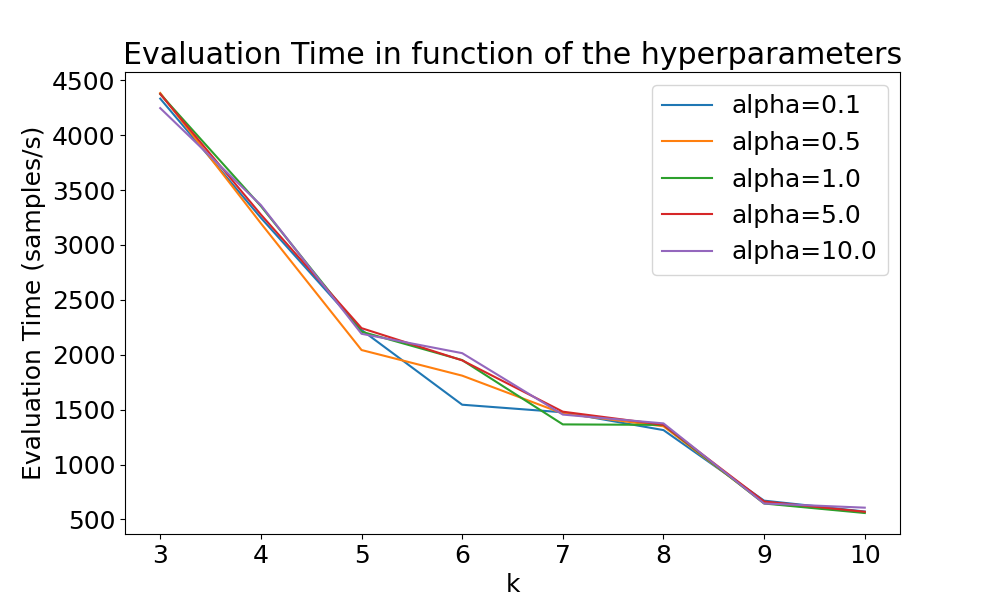}
        \captionsetup{width=.9\linewidth}
        \caption{AI-human-text (2181 chars/sample)}
        \label{fig:hyperparameters_time1}
    \end{subfigure}
    \caption{Time performance for the grid search of the hyperparameters $k$ and $\alpha$ for the datasets HC3 (On the left) and AI-human-text (On the right).}
    \label{fig:hyperparameters_time}
\end{figure}

Figure \ref{fig:hyperparameters} shows the F1 score and Fig. \ref{fig:hyperparameters_time} shows the time performance of the grid search for the hyperparameters $k$ and $\alpha$, using both datasets.

After analysing these results, we chose the hyperparameters $k=8$ and $\alpha=0.5$. The value of $k$ was selected because it consistently delivered a high F1 score without significant degradation in time performance, which started to appear for $k>8$. Similarly, $\alpha$ was chosen for its optimal balance, providing strong F1 scores across both datasets and different values of $k$, without affecting the algorithm's performance.

\subsection*{Alphabet Trimming}

To understand the trade-off between alphabet ($\Sigma$) size and classification impact, we performed a trimming experience. For this purpose, we used the previously optimised parameters and the following alphabets for calculating the F1 scores:
\begin{itemize}
    \item[] \textbf{$\Sigma_1$}: \verb|" abcdefghijklmnopqrstuvwxyz"|
    \item[] \textbf{$\Sigma_2$}: \verb|"1234567890 abcdefghijklmnopqrstuvwxyz"|
    \item[] \textbf{$\Sigma_3$}: \verb|"1234567890 abcdefghijklmnopqrstuvwxyz.,!?'\"\"/\\;:_-"|
    \item[] \textbf{$\Sigma_4$}: \verb|"1234567890 abcdefghijklmnopqrstuvwxyz.,!?'\"\"/\\;:_-@|\newline\verb|#$%^&*()[]{}<>"|
\end{itemize}

\begin{table}[H]
\centering
\caption{F1 scores using different alphabets for
the HC3 and AI-human-text datasets.}
\begin{tabular}{cc}
    \begin{subtable}[t]{0.45\textwidth}
        \centering
        \caption{HC3 dataset.}
        \vspace{0.3cm}
        \begin{tabular}{@{}lcc@{}}
        \toprule
        \textbf{Dataset}                        & \textbf{Alphabet} & \textbf{F1 score} \\ \midrule
        \multirow{4}{*}{HC3}           & {$\Sigma_1$}        & 0.9712        \\
                                       & {$\Sigma_2$}        & 0.9707        \\
                                       & {$\Sigma_3$}        & 0.9785        \\
                                       & {$\Sigma_4$}        & 0.9762        \\ \bottomrule
        \end{tabular}
    \end{subtable}
    &
    \begin{subtable}[t]{0.45\textwidth}
        \centering
        \caption{AI-human-text dataset.}
        \vspace{0.3cm}
        \begin{tabular}{@{}lcc@{}}
        \toprule
        \textbf{Dataset}                        & \textbf{Alphabet} & \textbf{F1 score} \\ \midrule
        \multirow{4}{*}{AI-human-text} & {$\Sigma_1$}        & 0.9965        \\
                                       & {$\Sigma_2$}        & 0.9964        \\
                                       & {$\Sigma_3$}        & 0.9963        \\
                                       & {$\Sigma_4$}        & 0.9961        \\ \bottomrule
        \end{tabular}
    \end{subtable}
\end{tabular}
\label{tab:different_alphabets}
\end{table}

From Table \ref{tab:different_alphabets}, we can see that the usage of different alphabets did not have a significant impact in the performance of the algorithm, neither in the same dataset nor between datasets. The F1 score varied from 0.9707 to 0.9785 in the first dataset and from 0.9961 to 0.9965 in the second dataset. In order to try to generate a more general model that could be used in different datasets without overfitting, we decided to use the \textbf{$\Sigma_2$} for both datasets as it was the one that averaged the best performance on both datasets.

\Section{Results}
Using the optimal parameters previously identified, we conducted three distinct analyses: first, to understand the impact of reference text length on classifier performance; second, to assess how the size of the target text influences the classifier's decisions; and third, to evaluate the overall performance (accuracy and computational resources) of the classifier on the datasets.

\subsection*{Influence of reference length}
\begin{figure}[H]
    \centering
    \begin{subfigure}[b]{0.49\linewidth}
        \includegraphics[width=\linewidth]{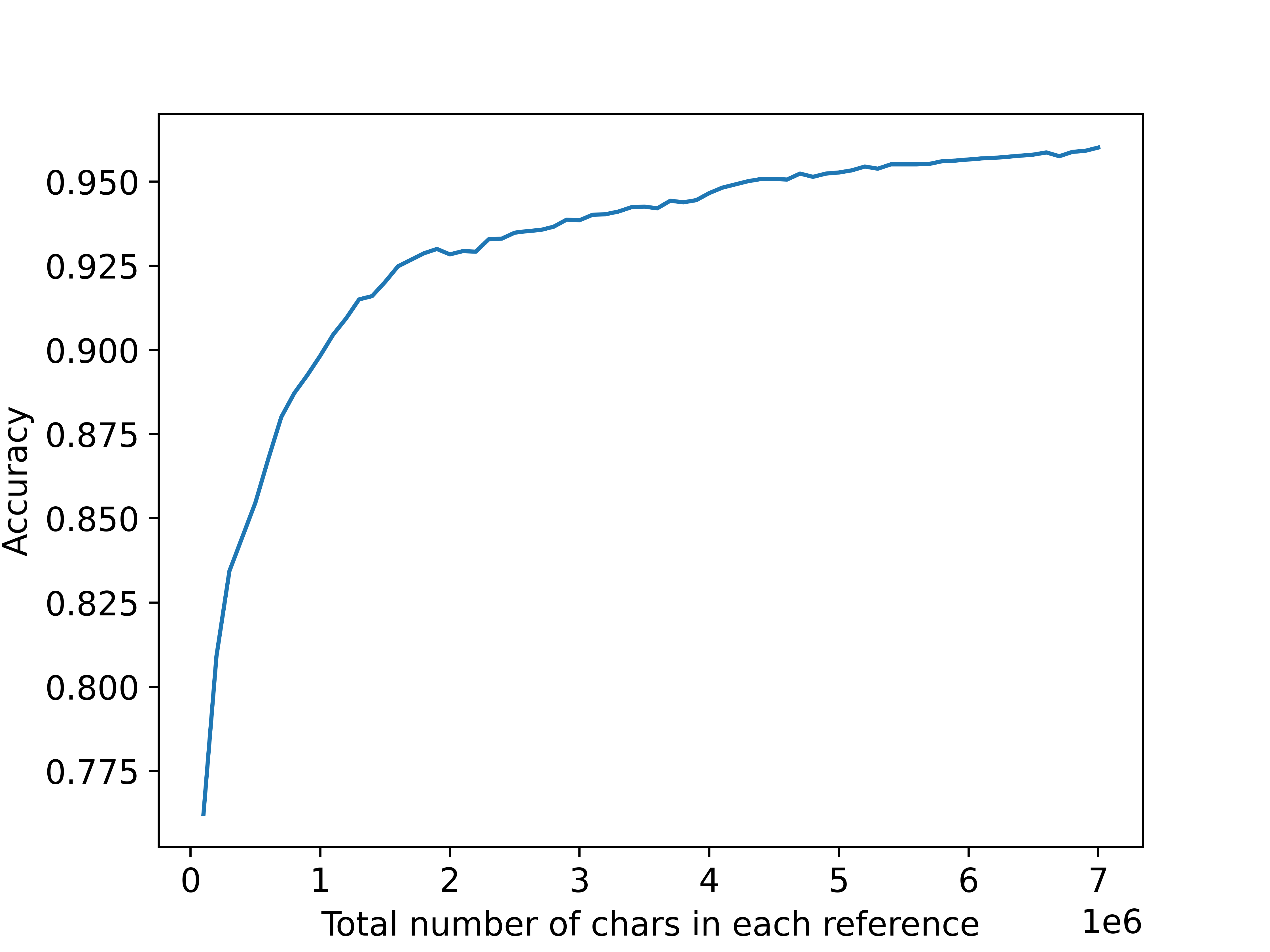}
        \captionsetup{width=.9\linewidth}
        \caption{HC3}
        \label{fig:references_data2}
    \end{subfigure}
    \begin{subfigure}[b]{0.49\linewidth}
        \includegraphics[width=\linewidth]{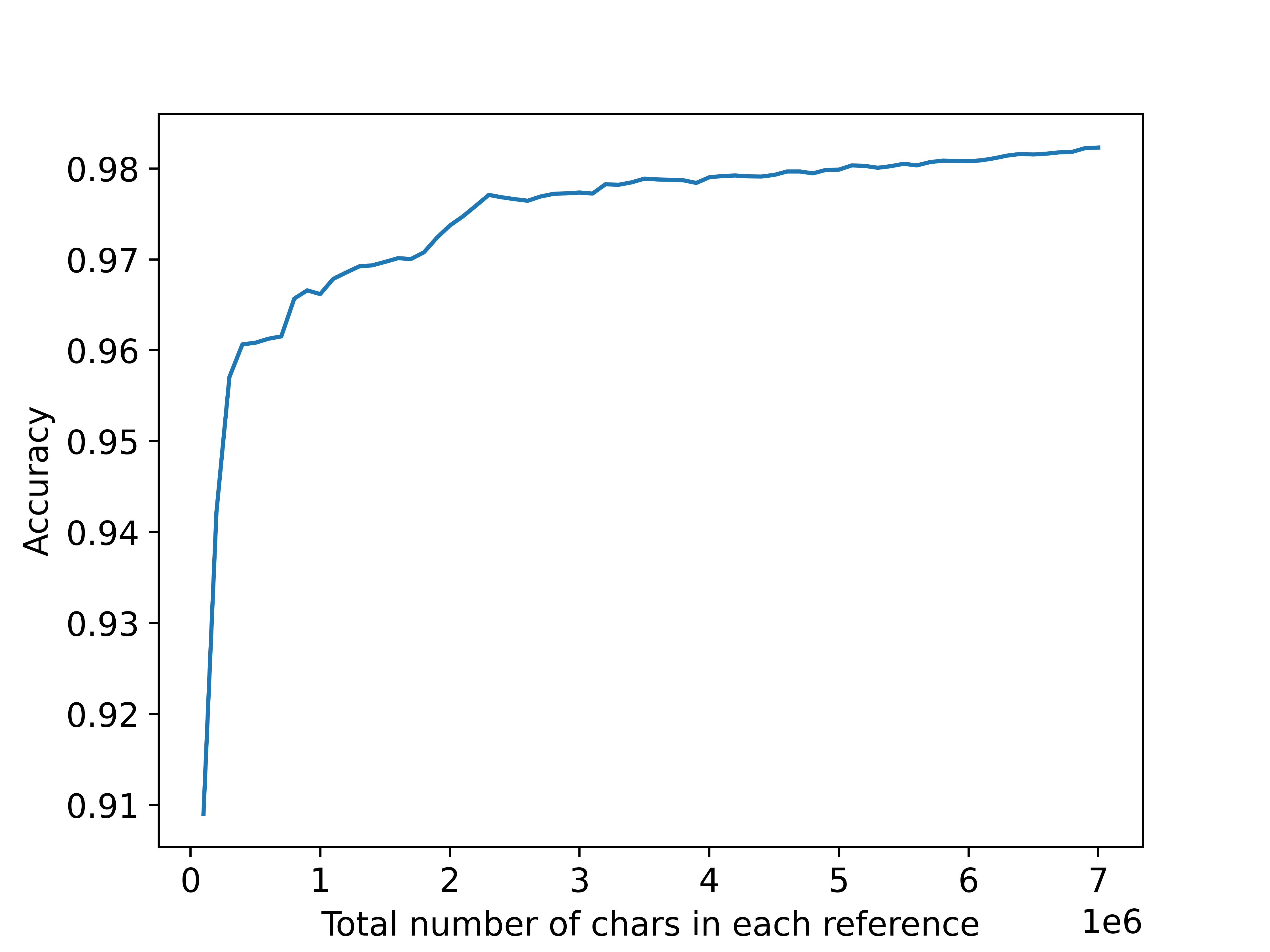}
        \captionsetup{width=.9\linewidth}
        \caption{AI-human-text}
        \label{fig:references_data1}
    \end{subfigure}
    \caption{Classifier performance evolution as reference text length increases.}
    \label{fig:references_length}
\end{figure}

We trained the models using different lengths for both references, where the length was measured by the total number of characters. The lengths considered were from 100,000 to 7 million characters, with increments of 100,000 characters. Then, we determined each classifier's accuracy on the test set. Figures \ref{fig:references_data2} and \ref{fig:references_data1} illustrate the results obtained for the HC3 and AI-human-text datasets, respectively.

From both datasets, it is clear that as the overall length of the reference texts increase, the classifiers’ performance also increases. This behaviour is expected, since the models have more information to learn from and can better distinguish between human and AI text. However, it is also important to note that the performance tends to increase at a slower pace after a certain length.

\subsection*{Influence of target sample}

\begin{figure}[H]
    \centering
    \begin{subfigure}[b]{0.49\linewidth}
        \includegraphics[width=\linewidth]{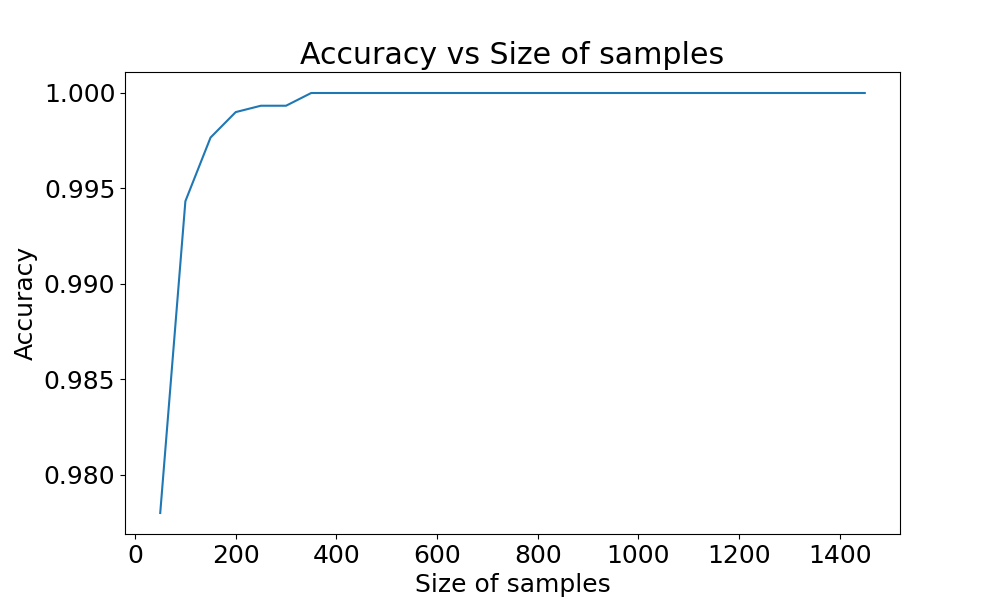}
        \captionsetup{width=.9\linewidth}
        \caption{HC3}
        \label{fig:target_sizes_data2}
    \end{subfigure}%
    \begin{subfigure}[b]{0.49\linewidth}
        \includegraphics[width=\linewidth]{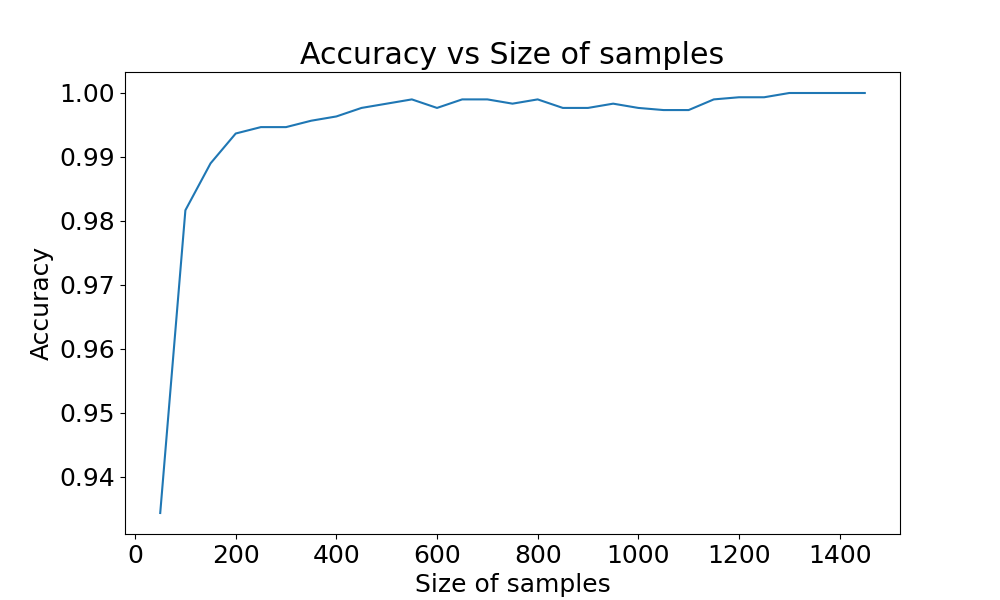}
        \captionsetup{width=.9\linewidth}
        \caption{AI-human-text}
        \label{fig:target_sizes_data1}
    \end{subfigure}
    \caption{Accuracy in function of the length of the target texts for the datasets HC3 (On the left) and AI-human-text (On the right).}
    \label{fig:target_sizes_data}
\end{figure}

Using the model trained with all the data, we analysed how the classifier behaves when the target text has different lengths. In this analysis, for each class, we used 1,500 samples of the test set with lengths of at least and near 1,500 characters. With these 3,000 samples, we ran the classification algorithm and calculated its accuracy, but only using the first $N$ characters of the target texts with increments of 50 characters for each iteration. The results obtained for each dataset are illustrated in Fig. \ref{fig:target_sizes_data}.

From the results in Fig. \ref{fig:target_sizes_data}, we can see that the classifier's performance keeps increasing as the number of characters of the target text also increases, tending to stabilise after a certain number of characters. For the case of the HC3 dataset, the algorithm's performance seems to stabilise its growth rate after 400 characters and, for the AI-human-text dataset, the performance seems to stabilise its growth rate after around 600 characters.

This behaviour shows that the classifier can distinguish better between human and AI samples when the target text has more information, allowing the model to better understand the patterns in the text.

\subsection*{Classifier testing}
\begin{figure}[ht]
    \centering
    \begin{subfigure}[b]{0.49\linewidth}
        \includegraphics[width=\linewidth]{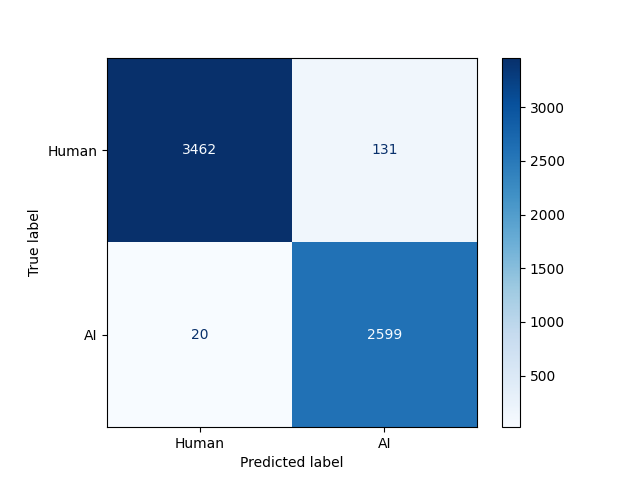}
        \caption{HC3}
        \label{fig:confusion_matrix2}
    \end{subfigure}%
    \begin{subfigure}[b]{0.49\linewidth}
        \includegraphics[width=\linewidth]{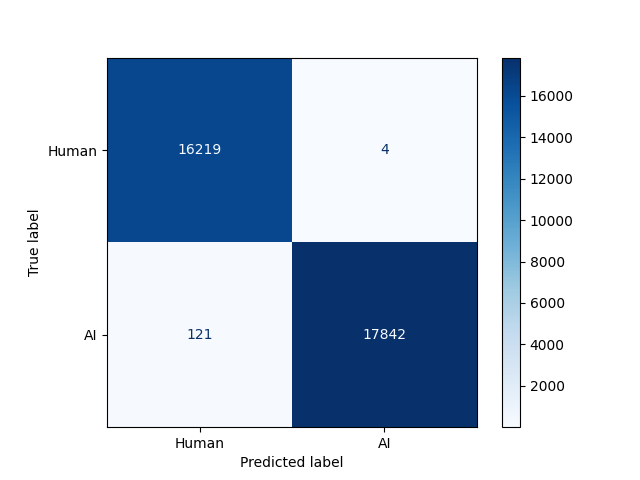}
        \caption{AI-human-text}
        \label{fig:confusion_matrix1}
    \end{subfigure}
    \caption{Confusion matrix for HC3 and AI-human-text datasets.}
    \label{fig:confusion_matrix}
\end{figure}

To test the classifier's overall performance, we tested it against the test set of each dataset. The results obtained for each dataset are illustrated in the confusion matrices in Fig. \ref{fig:confusion_matrix}.

From the results in Fig. \ref{fig:confusion_matrix}, we can see that the classifier performed very well and had low quantities of false positives and false negatives for each class. This is an excellent result since it shows that the target texts are being classified with extremely high accuracy and F1 score.

Table \ref{tab:results} shows the metrics results using the algorithm for each dataset. The classifier's performance was very good, outperforming our expectations.

\begin{table}[H]
    \centering
    \caption{Results of the classifier for each dataset.}
    \begin{tabular}{@{}lcc@{}}
        \toprule
        \textbf{Dataset} & \textbf{Accuracy} & \textbf{F1 score} \\ \midrule
        HC3 & 0.9757 & 0.9752 \\
        AI-human-text & 0.9963 & 0.9963 \\ \bottomrule
    \end{tabular}
    \label{tab:results}
\end{table}

\subsection*{Performance}

Besides evaluating the classifier's score on the test sets, we also analysed the performance of the algorithm in terms of time and RAM usage. The results obtained for each dataset are illustrated in Tables \ref{tab:creation_metrics} and \ref{tab:infer_metrics}. The experiments were conducted on a single machine with an Intel Core i7-13700KF CPU and 32GB of RAM.

\begin{table}[H]
    \centering
    \caption{Metrics for the creation of the models for each dataset.}
    \begin{tabular}{@{}lrc@{}}
        \toprule
        \textbf{Dataset} & \textbf{Time (s)} & \textbf{RAM (MB)} \\ \midrule
        HC3 & 22.02 & 1,050 \\
        AI-human-text & 154.38 & 2,157 \\ \bottomrule
    \end{tabular}
    \label{tab:creation_metrics}
\end{table}

\begin{table}[H]
    \centering
    \caption{Metrics for the inference of the models for each dataset.}
    \begin{tabular}{@{}lcccc@{}}
        \toprule
        \textbf{Dataset} & \textbf{RAM (MB)} & \textbf{Samples/s} & \textbf{Chars/sample} & \textbf{Chars/s} \\ \midrule
        HC3 & 1,664 & 3,620 & 858 & 3.10M \\
        AI-human-text & 4,039 & 1,350 & 2,181 & 2.95M \\ \bottomrule
    \end{tabular}
    \label{tab:infer_metrics}
\end{table}

The results show that the algorithm scales well with the size of the dataset in terms of time and RAM usage, which can be explained by the fact that new table entries are created less frequently as the dataset grows. 
The higher RAM usage for inference is due to the need to store two models in memory, one for each class. Despite being a single-threaded algorithm, the time performance is good, with the algorithm being able to process around 3.1M characters per second for the HC3 dataset and 2.95M characters per second for the AI-human-text dataset. Moreover, only single-core processing was used, so significant improvements in computational time are possible with parallel computing.

For both datasets, most of the models proposed by the community are based on large language models, more specifically the BERT \cite{devlin2019bertpretrainingdeepbidirectional} model. To train this model it was necessary 3.3B words, and it took approximately 4 days to train it on 64 TPUs. Putting this model and our algorithm side by side, it is clear that our algorithm is much more efficient in terms of time and resources, making it a good alternative for scenarios where computational resources are limited.

\Section{Conclusions}

In this paper, we introduced AIDetx, a method for distinguishing between human- and AI-rewritten text using finite-context models. AIDetx builds separate models for human and AI text, classifying target texts via relative data compression.

We optimized performance through hyperparameter tuning and tested various alphabets to enhance classification. Additionally, we assessed accuracy across different reference and target text lengths.

In the testing phase, AIDetx achieved 97\% accuracy on the HC3 dataset and over 99\% on the AI-human-text dataset, demonstrating the potential of data compression for distinguishing between human- and AI-rewritten text. While performance may vary on other datasets, AIDetx offers greater transparency, requires substantial less computational power, and is more interpretable than other machine learning methods.

\section*{Acknowledgements}

We acknowledge national funds through FCT – Fundação para a Ciência e a Tecnologia, I.P., in the context of the projects CEECINST/00026/2018 and UIDB/00127/2020 (\url{
https://doi.org/10.54499/UIDB/00127/2020}).

\Section{References}
\bibliographystyle{IEEEbib}


\end{document}